\documentclass[runningheads]{llncs}
\usepackage{graphicx}
\usepackage{comment}
\usepackage{amsmath,amssymb} 
\usepackage{color}
\usepackage{adjustbox}
\usepackage{subfig}
\usepackage{algorithm}
\usepackage{algorithmic}
\usepackage{multirow}
\usepackage{multicol}
\usepackage{enumerate}
\usepackage{verbatim}

\newcommand{\ie}{\textit{i}.\textit{e}., }

\newcommand{\tabincell}[2]{\begin{tabular}{@{}#1@{}}#2\end{tabular}}
\begin{document}
\title{Quality Control of Neuron Reconstruction Based on Deep Learning}
\titlerunning{Quality Control of Neuron Reconstruction Based on Deep Learning} 
\authorrunning{Donghuan Lu et al.} 
\author{Donghuan Lu\inst{1}, Sujun Zhao\inst{2}, Peng Xie\inst{2}, Kai Ma\inst{1}, Lijuan Liu\inst{2}, Yefeng Zheng\inst{1}}
\institute{Tencent \and Southeast University}

%
%
\maketitle              
\begin{abstract}
Neuron reconstruction is essential to generate exquisite neuron connectivity map for understanding brain function. Despite the significant amount of effect that has been made on automatic reconstruction methods, manual tracing by well-trained human annotators is still necessary. To ensure the quality of reconstructed neurons and provide guidance for annotators to improve their efficiency, we propose a deep learning based quality control method for neuron reconstruction in this paper. By formulating the quality control problem into a binary classification task regarding each single point, the proposed approach overcomes the technical difficulties resulting from the large image size and complex neuron morphology. Not only it provides the evaluation of reconstruction quality, but also can locate exactly where the wrong tracing begins. This work presents one of the first comprehensive studies for whole-brain scale quality control of neuron reconstructions. Experiments on five-fold cross validation with a large dataset demonstrate that the proposed approach can detect 74.7\% errors with only 1.4\% false alerts.

\keywords{Neuron reconstruction \and Quality control \and Deep learning.}
\end{abstract}
\section{Introduction}
\begin{figure}[th]
	\centering
	  \subfloat[]{\label{fig:whole_brain}%
		\includegraphics[width=.45\linewidth, height=.4\linewidth]{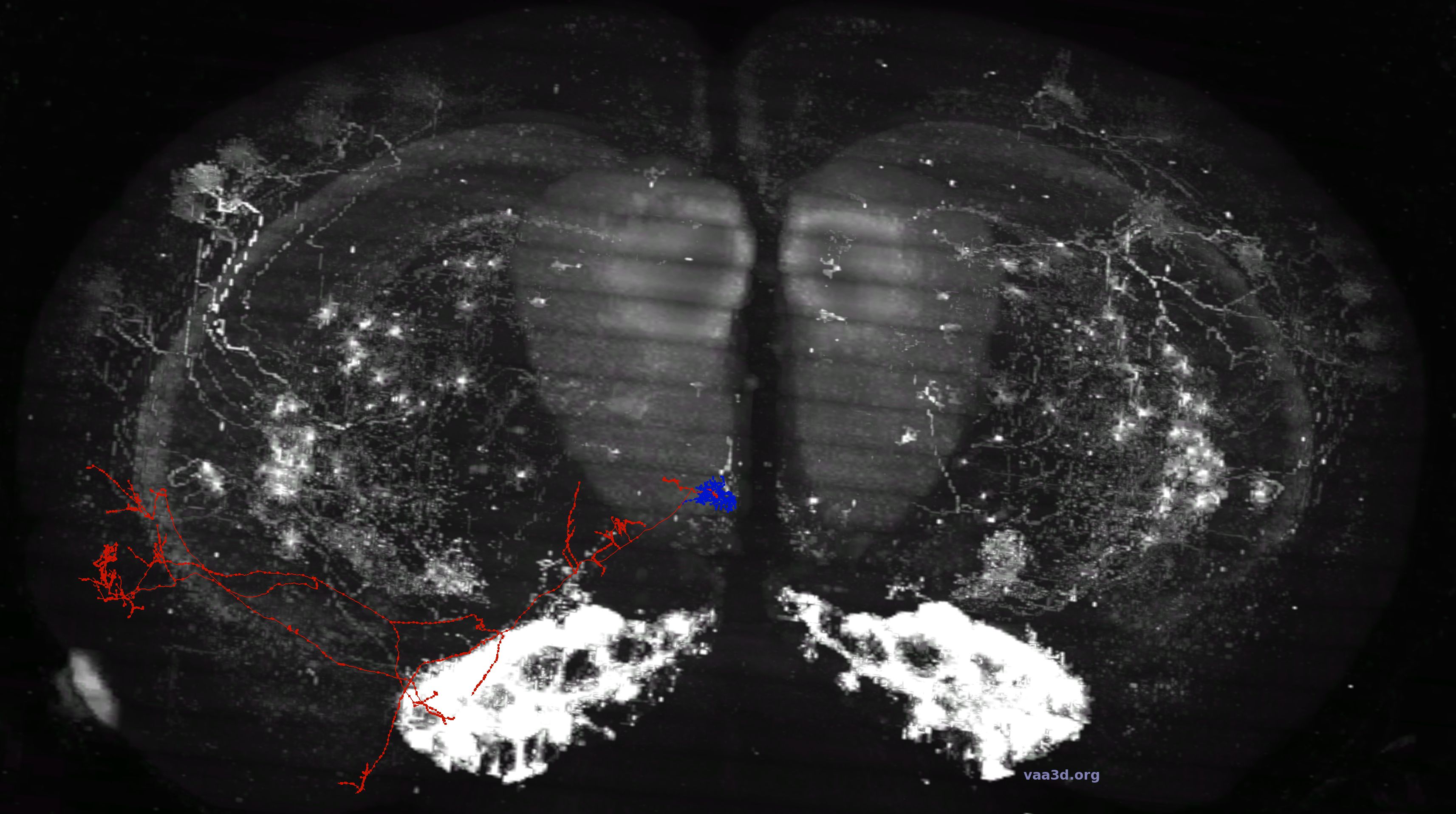}}
	  \hspace{0.05\linewidth}
	  \subfloat[]{\label{fig:neuron_shape}%
		\includegraphics[width=.45\linewidth, height=.4\linewidth]{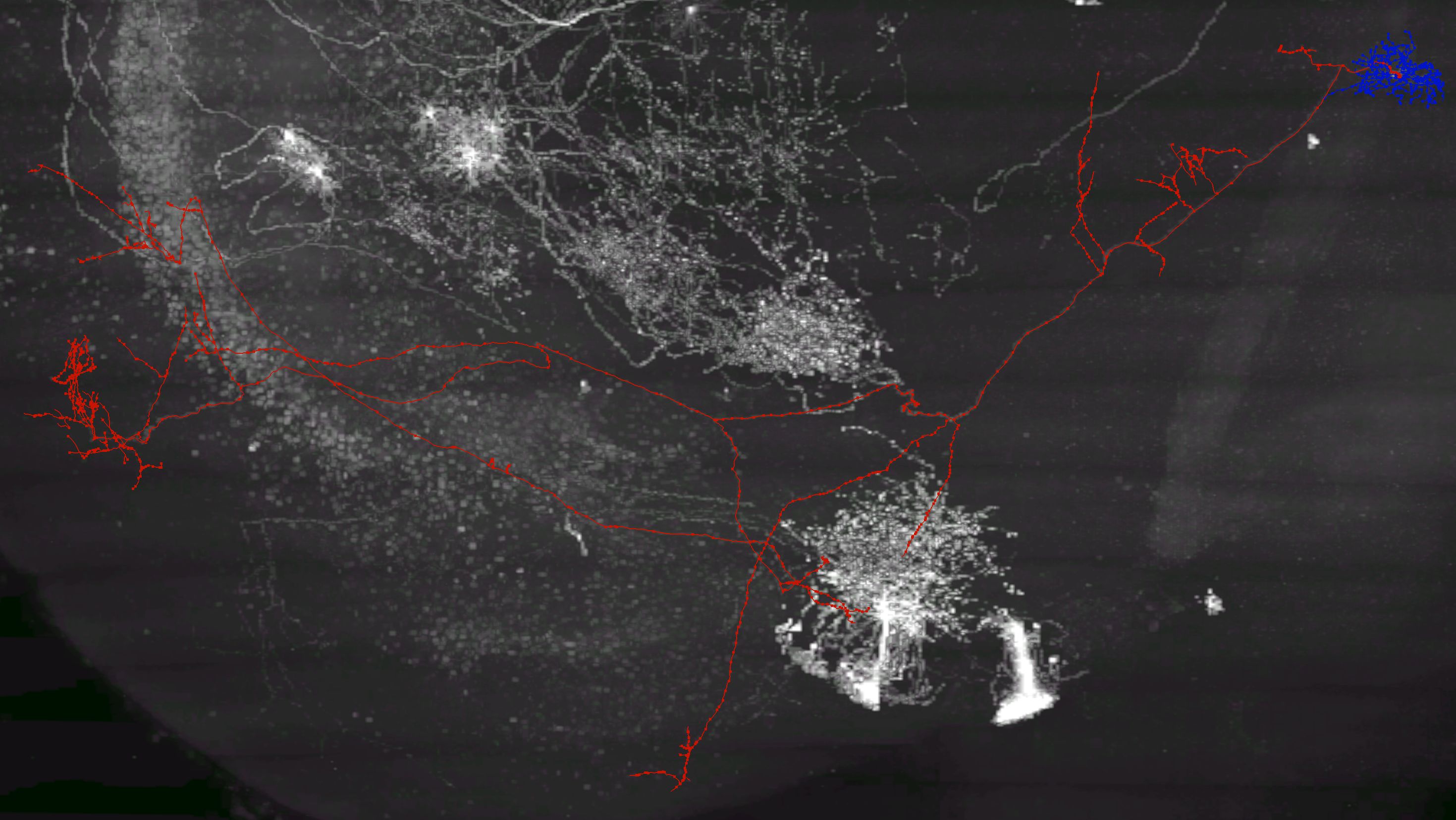}}
	  \caption{Example of a reconstructed neuron in a mouse brain. The visualization in the whole brain is displayed in (a), while the enlarged visualization is shown in (b). Red represents the axon, and blue stands for the dendrites.}
	  \label{fig:NeuronExample}
\end{figure}

Neuronal connectivity is one of the key topics on the frontier of brain science~\cite{sporns2005human}. The exquisite connectivity map is fundamental to understand human intelligence and emotion, and beneficial for designing frameworks of artificial intelligence algorithms. Reconstruction of full morphology for every single neuron provides the ultimate resolution in mapping connectivity at whole brain level~\cite{winnubst2019reconstruction,wang2019complete}. Specifically, neurons are genetically labelled and imaged, followed by digital reconstructions to represent the spatial location and topology as a tree-like structure. The visualization of a reconstructed neuron in the whole mouse brain using the Vaa3D platform~\cite{peng2010v3d} is displayed in Fig.~\ref{fig:NeuronExample}.

Neuron reconstruction, or tracing, is a challenging task as axonal/dendritic arborization can be both dense and complex. Reconstruction of the axonal arbor of one single typical neuron requires tracing dozens of branches in a 3D image with thousands of voxels in each dimension. In addition, imaging signals can be weak in certain brain areas and neighboring neurons often have branches that are close to each other, as shown in Fig.~\ref{fig:two_close_neurons}, making it difficult to discriminate the connections. As a result, although dozens of methods~\cite{liu2016rivulet,wan2016automatic,wang2017ensemble,wang2018automatic,yang2019fmst} have been proposed for the purpose of automatic neuron tracing, their results are still far from satisfactory. Manual tracing by well-trained human annotators is still an indispensable step, where integrity of tracing results is controlled by either consensus from multiple annotators~\cite{winnubst2019reconstruction,wang2019complete} or by several rounds of examinations (unpublished intermediate annotations).

The disadvantage of manual tracing is that it cannot be systematic and is highly labor-intensive and time-consuming. Due to the complexity of neuron morphology, even well-trained annotators may need a Virtual Reality (VR) equipment for correct reconstruction~\cite{wang2019teravr}. Therefore, it is important to introduce systematic algorithms to improve the quality and efficiency. To this end, we collaborate with neuron annotators and find out that the identification of key points, such as branching points or termination points, is an essential step in tracing, as missing or erroneously adding of such a single point may lead to serious topological errors. The identification of these points is also one of the most time-consuming steps in the process of manual tracing. Therefore, the neuron annotators can benefit a lot from a computer-aided system which can provide guidance to determine these key points.

In this work, we propose a framework to formulate the quality control problem of neuron reconstruction into a binary classification task for point of interest (where the wrong tracing starts). Benefiting from the recent development of deep learning~\cite{simonyan2014very,he2016deep,litjens2017survey,huang2017densely,zhang2018survey}, several commonly used networks for 2D image recognition are converted into 3D version and their performance on this problem is investigated. The cross validation experiments on a large dataset demonstrate that the proposed approach can not only evaluate the quality of neuron reconstruction, but also provide guidance for the annotators to locate the problem without too many false alerts.


\section{Materials and Methods}
\subsection{Data}
The manual reconstruction (or tracing) of neuron morphology consists of several rounds of examination. In each round, an annotator manually traces all the neuronal points, and then another annotator validates the reconstructed results and marks the mistakes, which are going to be corrected in the next round. Such process usually needs to be repeated for three times unless there are no further mistakes. The neurons passing the final round are used as the gold standard for the correct reconstructions, while the neurons with marked mistakes can be employed as the incorrect cases. 

In this study, there are 254 neurons with correct reconstructions, and each of them has 1 or 2 wrong reconstructions from different rounds (421 wrong reconstructions in total). To clarify notations, up case letter $R$ is used to represent a neuron reconstruction and lower case letter $p$ is applied to indicate a single neuronal point. A reconstructed neuron $R_a$ is stored in an SWC file, which is a standardized neuromorphometric format~\cite{o2020module} and commonly used for neuron reconstruction sharing as well as neuronal morphology analysis. Each line in an SWC file represents a neuronal point $p_1$ with its seven properties, including the voxel’s identifier number, neuronal type (soma, axon and dendrite), x coordinate, y coordinate, z coordinate, radius, and the identifier number of its parent $p_2$ ($p_1$ is known as the child of $p_2$). In addition, the corresponding 3D optical microscope images cropped based on the SWC files are used to provide the intensity information of the neurons and the surrounding background. 

\subsection{Problem Formulation}
\label{sec:preprocess}
To determine whether the reconstruction of a neuron is correct or not, an intuitive way is to first generate a 3D binary map based on the SWC file, in which the intensities of the voxels labeled as neuronal points are set to be 1 and the intensities of the rest voxels are set to be 0; then add this binary map as another channel to the corresponding optical microscope image; and finally feed the concatenated image into a deep neural network for classification. However, the size of the 3D optical microscope image could be extremely large, \ie thousands of voxels in each dimension. With such a large image as input, the classifier requires several terabytes of memory and months to train. In addition, due to the sparsity of a single neuron reconstruction as shown in Fig.~\ref{fig:NeuronExample}, this large image may feed too much irrelevant information to the network and lead to inferior performance. On the other hand, such a classifier can only provide the evaluation of the whole neuron. It would still be difficult for the annotators to locate the problem due to the tens of thousands points in each neuron.  

Therefore, we choose to perform the classification on each single point in this work instead of the whole image. A commonly used approach for point-wise recognition is to crop a small region around a point from the whole image and feed it to a deep neural network to determine the category of this point. However, an incorrect reconstruction does not mean that all its neuronal points are wrongly traced. Comparison between the correct and wrong reconstruction is necessary to determine the category of points.

Considering the possible disturbance of manual tracing, slight deviations of coordinates should be allowed. We define a point $p_i$ from reconstruction $R_a$ having a match in reconstruction $R_b$ if 1) $R_a$ and $R_b$ are the reconstructions of the same neuron; 2) there is a point $p_j$ in $R_b$ whose Euclidean distance to $p_i$ is less than a threshold (set as 4 in this study), \ie 
\begin{equation}
\label{eq:eculidean_distance}
\sqrt{(p_{i,x}-p_{j,x})^2+(p_{i,y}-p_{j,y})^2+(p_{i,z}-p_{j,z})^2} < 4
\end{equation}
where $p_{i,x}$, $p_{i,y}$, and $p_{i,z}$ represent the 3D coordinates of point $p_i$. 

\begin{figure}[th]
	\centering
	\subfloat[]{\label{fig:close_neuron1}%
		\includegraphics[width=.45\linewidth, height=.45\linewidth]{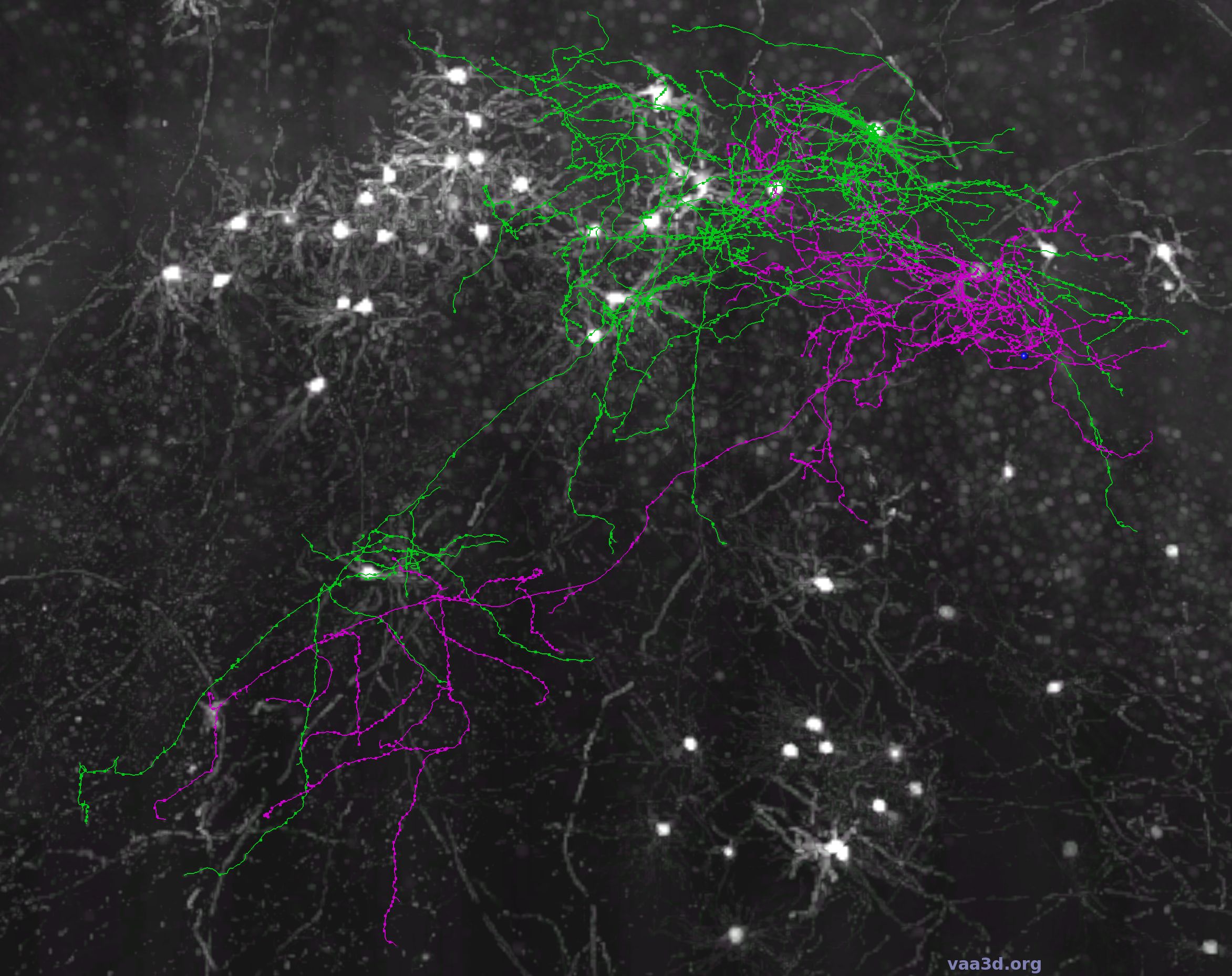}}
	  \hspace{0.05\linewidth}
	  \subfloat[]{\label{fig:close_neuron2}%
		\includegraphics[width=.45\linewidth, height=.45\linewidth]{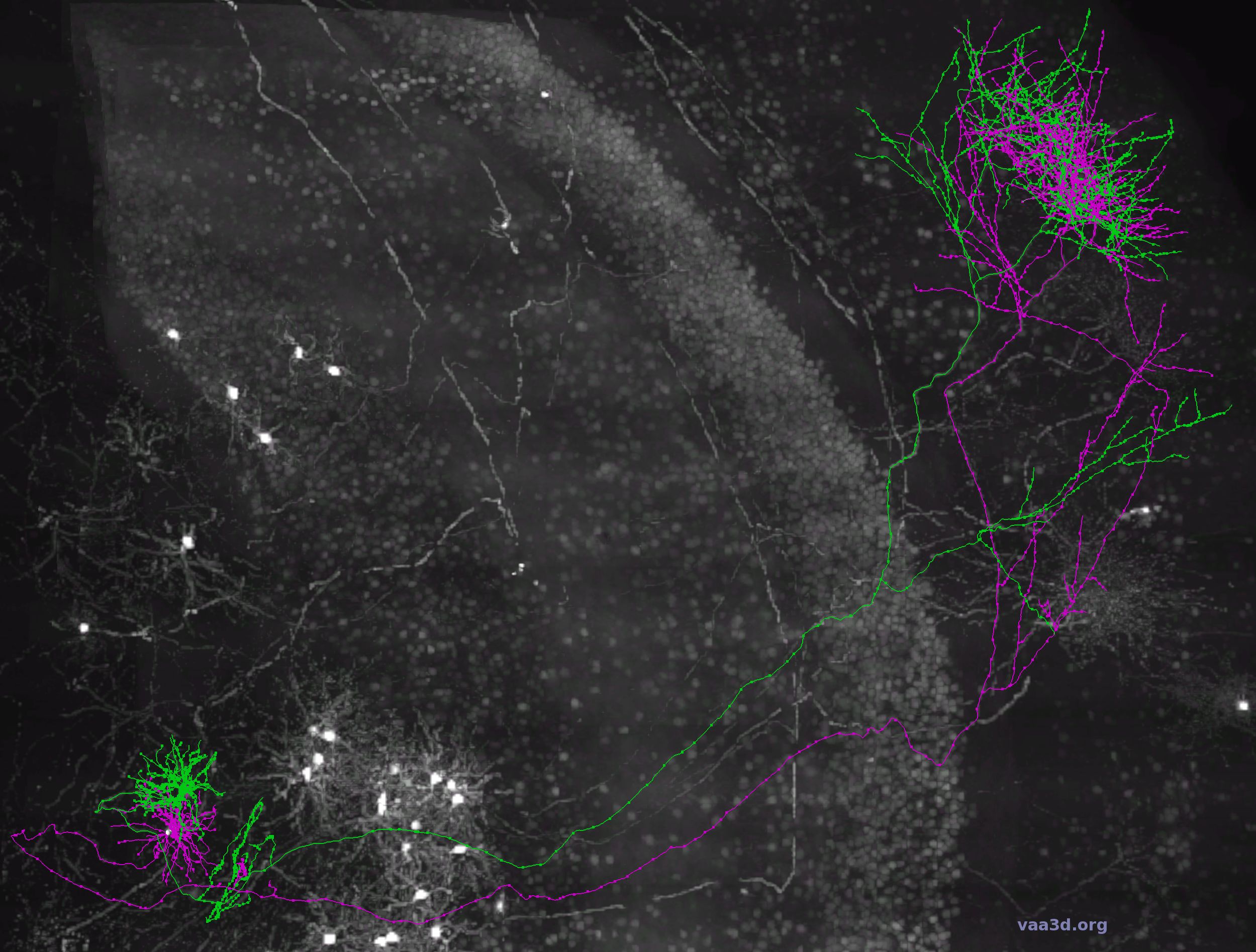}}
	\caption{Visualization of two neurons close to each other. Green and purple represent points belong to different neurons. Note how close two neurons could be.}
	\label{fig:two_close_neurons}
\end{figure}

Although $p_i$ should be labelled as a wrongly traced point if $p_i$ does not have any match in the correct reconstruction of the same neuron, it does not necessarily mean that $p_i$ belongs to background. As shown in Fig.~\ref{fig:two_close_neurons}, two neurons could be so close that the annotator accidentally jumps from one neuron to the other, which should still be considered as wrong tracing. However, such a mistake is theoretically impossible to detect with only a small neighbourhood for most points in that branch because it is actually a correct reconstruction for the other neuron. Therefore, to determine whether the reconstruction of a point is correct or not may require tracing back to hundreds of points ahead, which results in cropping a large image and may lead to inferior performance due to the interference of the redundant regions. 

To resolve this issue, we propose to only detect the points which initialize the wrong tracing instead of all the points that do not belong to the neuron and the points that have been missed during the reconstruction. A point $p_i$ from the wrong reconstruction $R_a$ is defined as a point where the wrong tracing begins under two conditions: 1) point $p_i$ must have a match $p_j$ in the correct reconstruction $R_b$; 2) there is a child of point $p_i$ who does not have any match in $R_b$, \ie a point has been wrongly traced, or there is a child of point $p_j$ does not have any match in $R_a$, \ie a point has been missed. Therefore, the quality control problem of neuron reconstruction is converted to a binary classification between $p_i$ (where the wrongly traced begins, denoted as POI, \ie point of interest, for the rest of paper) and $p_j$ (the match point of $p_i$) based on their neighbourhoods in the reconstruction $R_b$ and $R_a$, respectively. It is worth mentioning that although such a strategy cannot provide the location of all the wrongly traced points, it is enough or even better for the annotators than just offering the location of POIs. In order to identify and fix the problem, the annotators also need to locate where the wrong tracing begins, and they can easily remove all the following points from there if the reconstruction jumps to another neuron or leaks into background tissues.

For the rest of paper, the labels of POIs are set as 1 to represent the experimental group, while their match points are labelled as 0 indicating that these samples belong to the control group. Although theoretically, every point in the correct reconstructions could be used as a sample of the control group, using all the correct points will lead to a highly unbalanced training set (6,423 vs. 20,548,189) and cost much more time for training. Instead we only use the ones whose match points are POIs as the constant control group, and randomly add a few other correct points for training in each iteration.

\subsection{Network Architectures}
With the recent development of deep learning, many networks have been proposed for 2D or 3D image recognition~\cite{simonyan2014very,he2016deep,huang2017densely,song2016deep,griffiths2019review}. However, the concept of 3D image in medical imaging community is not the same as in natural image field, whose 3D images are mostly composed of a point cloud, a 2D image with an additional channel for depth information, or video with time as the third dimension. Each sample used in this study is the concatenation of an optical microscope image and a binary image. It is in the format of a 3D image with three spatial coordinates and two channels, and saved as a 4D matrix. Therefore, most 3D deep learning approaches can barely be used for this problem. Instead, we select six network architectures which have delivered state-of-the-art performance for many 2D image recognition problems, and convert them into 3D versions, including VGG11 and VGG16~\cite{simonyan2014very}, ResNet101 and ResNet152~\cite{he2016deep}, DenseNet121 and DenseNet201~\cite{huang2017densely}. The exact configurations of these networks are shown as in Table~\ref{table:network_configuration}. Note that because the size of the cropped image for each point is only $32\times 32\times 32$ voxels, the strides of the first convolutional layers in ResNet101, ResNet152, DenseNet121 and DenseNet201 are set to one to retain sufficient detail information for the rest of layers. In addition, the last convolutional layer with a stride of two has been removed from ResNet101 and ResNet152 to ensure the size of input for the last pooling layer is larger than one. 

\begin{table}[th]
    \caption{Network architectures used in this study. $conv N$, $maxpool N$ and $avgpool N$ represents 3D convolution, max pooling and average pooling with $N\times N\times N$ kernel, respectively. $fc$ stands for a fully connected layer. The stride for convolution is 1 unless otherwise stated, and the stride for max pooling and average pooling are both 2.}
    \centering
    \setlength{\tabcolsep}{3.5pt}
    \begin{tabular}{|c|c|c|c|c|c|}
		\hline
		VGG11 & VGG16 & ResNet101 & ResNet152 & DenseNet121 & DenseNet201 \\
		\hline
		$conv 3$ & $\left[conv 3\right]\times 2$ & \multicolumn{4}{|c|}{$conv 7$} \\
		\hline
		\multicolumn{2}{|c|}{$maxpool 2$} & \multicolumn{4}{|c|}{$maxpool 3$}\\
		\hline 
		\multirow{2}*{\tabincell{c}{\\$conv 3$}} & \multirow{2}*{\tabincell{c}{\\$\left[conv 3\right]\times 2$}} & \multirow{2}{*}{$\left[\begin{matrix}
		conv 1 \\
		conv 3 \\
		conv 1
		\end{matrix}\right]\times 3$} & \multirow{2}{*}{$\left[\begin{matrix}
		conv 1 \\
		conv 3 \\
		conv 1
		\end{matrix}\right]\times 3$} & \multirow{1}{*}{$\left[\begin{matrix}
		conv 1 \\
		conv 3
		\end{matrix}\right]\times 6$} & \multirow{1}{*}{$\left[\begin{matrix}
		conv 1 \\
		conv 3
		\end{matrix}\right]\times 6$}\\[11.5pt]
		\cline{5-6}
		& & & & \multicolumn{2}{|c|}{$conv1$}\\
		\hline
		\multicolumn{2}{|c|}{$maxpool 2$} &\multicolumn{2}{|c|}{$conv1$, $stride$ 2} &\multicolumn{2}{|c|}{$avgpool 2$}\\
		\hline 
		\multirow{2}*{\tabincell{c}{\\$\left[conv 3\right]\times 2$}} & \multirow{2}*{\tabincell{c}{\\$\left[conv 3\right]\times 3$}} & \multirow{2}{*}{$\left[\begin{matrix}
		conv 1 \\
		conv 3 \\
		conv 1
		\end{matrix}\right]\times 4$} & \multirow{2}{*}{$\left[\begin{matrix}
		conv 1 \\
		conv 3 \\
		conv 1
		\end{matrix}\right]\times 8$} & \multirow{1}{*}{$\left[\begin{matrix}
		conv 1 \\
		conv 3
		\end{matrix}\right]\times 12$} & \multirow{1}{*}{$\left[\begin{matrix}
		conv 1 \\
		conv 3
		\end{matrix}\right]\times 12$}\\[11.5pt]
		\cline{5-6}
		& & & & \multicolumn{2}{|c|}{$conv1$}\\
		\hline
		\multicolumn{2}{|c|}{$maxpool 2$} &\multicolumn{2}{|c|}{$conv1$, $stride$ 2} &\multicolumn{2}{|c|}{$avgpool 2$}\\
		\hline 
		\multirow{2}*{\tabincell{c}{\\$\left[conv 3\right]\times 2$}} & \multirow{2}*{\tabincell{c}{\\$\left[conv 3\right]\times 3$}} & \multirow{2}{*}{$\left[\begin{matrix}
		conv 1 \\
		conv 3 \\
		conv 1
		\end{matrix}\right]\times 23$} & \multirow{2}{*}{$\left[\begin{matrix}
		conv 1 \\
		conv 3 \\
		conv 1
		\end{matrix}\right]\times 36$} & \multirow{1}{*}{$\left[\begin{matrix}
		conv 1 \\
		conv 3
		\end{matrix}\right]\times 24$} & \multirow{1}{*}{$\left[\begin{matrix}
		conv 1 \\
		conv 3
		\end{matrix}\right]\times 48$}\\[11.5pt]
		\cline{5-6}
		& & & & \multicolumn{2}{|c|}{$conv1$}\\
		\hline
		\multicolumn{2}{|c|}{$maxpool 2$} &\multicolumn{2}{|c|}{$conv1$, $stride$ 2} &\multicolumn{2}{|c|}{$avgpool 3$}\\
		\hline 
		\multirow{2}*{\tabincell{c}{\\$\left[conv 3\right]\times 2$}} & \multirow{2}*{\tabincell{c}{\\$\left[conv 3\right]\times 3$}} & \multirow{2}{*}{$\left[\begin{matrix}
		conv 1 \\
		conv 3 \\
		conv 1
		\end{matrix}\right]\times 3$} & \multirow{2}{*}{$\left[\begin{matrix}
		conv 1 \\
		conv 3 \\
		conv 1
		\end{matrix}\right]\times 3$} & \multirow{1}{*}{$\left[\begin{matrix}
		conv 1 \\
		conv 3
		\end{matrix}\right]\times 16$} & \multirow{1}{*}{$\left[\begin{matrix}
		conv 1 \\
		conv 3
		\end{matrix}\right]\times 32$}\\[11.5pt]
		\cline{5-6}
		& & & & \multicolumn{2}{|c|}{$conv1$}\\
		\hline
		\multicolumn{2}{|c|}{$maxpool 2$} &\multicolumn{2}{|c|}{$avgpool 2$} &\multicolumn{2}{|c|}{$avgpool 2$}\\
		\hline
		\multicolumn{2}{|c|}{$\left[fc\right]\times 3$, $softmax$} &\multicolumn{4}{|c|}{$fc$, $softmax$}\\
		\hline
	\end{tabular}
	\label{table:network_configuration}
\end{table}

\section{Experiments}
\subsection{Experimental Setup}
The networks are implemented with PyTorch~\cite{paszke2017automatic} on NVIDIA Tesla P40 GPUs. The parameters of all the networks are randomly initialized without any pretraining. Weighted cross entropy is applied as the loss function, and weight for each group is in inverse proportion to its sample number. The Adam~\cite{kingma2014adam} optimizer with $\beta_1=0.9$, $\beta_2=0.999$ is used for optimization without any weight decay. The learning rate starts with $1\times 10^{-5}$ and decreases to one-tenth after every 10 epochs. The batch size is set to 15 and the maximum number of epochs is set to 50. Five commonly used metrics are presented to evaluate the performance of the proposed framework, including area under the curve (AUC), accuracy, sensitivity, specificity and precision. These metrics range in $[0, 1]$, and a higher score implies better performance.

First, a five-fold cross validation experiment is performed to evaluate the proposed framework. Following the strategy stated in Section~\ref{sec:preprocess}, the points of interest from the wrong reconstructions and their match points in the correct reconstructions are used for cross validation, which are 6,423 pair of images extracted from 675 reconstructions of 254 neurons. To avoid potential bias, the images are randomly split into five folds on the neuron level, \ie the images from different reconstructions of the same neuron belong to the same fold and are used either all for training or all for test. Therefore, the number of points in each training set may not be identical. In each iteration during the training process, five other points are randomly selected from the correct reconstructions of the neurons belonging to the training sets. These points are also used for optimization to prevent potential overfitting, because the errors annotators make could concentrate on some special regions and training with only the POIs and their match points may lead to a network with inferior performance on other regions. Hence, each batch for training contains five POIs, their match points and five randomly selected points. Considering the number of points belonging to the control group is twice as many as the points belong to the experimental group, the weight is set as 1 for the control group and 2 for the experimental group.

To further evaluate the generalization ability of the framework, all the other points from the correct reconstructions (besides the match points of POIs) in the test sets are also utilized for evaluation with the network which has the best performance in the cross validation experiment. Note that other points of the wrong reconstruction cannot be used here because the label of most wrongly traced points should be set neither as 1 (POI, where the wrong tracing begins) nor as 0 (correctly traced) based on the current strategy. Therefore, all the points used in this experiment should belong to the control group, the accuracy is equivalent to the specificity.

\subsection{Main Result}
The results with different network architectures are shown as in Table~\ref{table:cross_validation}. Note that although some other random points are used for training, we only perform the evaluation for the classification of POIs and their match points here. The proposed framework with VGG11 as the classifier achieves the best performance, with an average AUC score of 94.9.0\% and an average accuracy of 86.6\%. The high sensitivity (74.7\%) and specificity (98.6\%) indicates that the 3D VGG11 network can detect most POIs with only a few false alerts. It demonstrates the capability that the proposed framework has to provide guidance for the annotators to locate where the wrong tracing begins. 

\begin{table}[th]
\caption{The results of the five-fold cross validation experiments with different network architectures. In each cell, the first number represents the average measurement, and the second number indicates the standard deviation.
         \label{table:cross_validation}}
\centering
\begin{tabular}{|c|c|c|c|c|c|}
\hline
Network       & AUC (\%)  & Accuracy (\%) & Sensitivity (\%) & Specificity (\%) & Precision (\%) \\ \hline
ResNet101~\cite{he2016deep}   & 84.8$\pm$5.1 & 79.3$\pm$3.5    & 71.3$\pm$9.4        & 87.3$\pm$3.9        & 87.9$\pm$3.8      \\ \hline
ResNet152~\cite{he2016deep}   & 86.7$\pm$5.3 & 81.8$\pm$3.6    & 71.1$\pm$8.7        & 92.4$\pm$1.9        & 90.5$\pm$1.3      \\ \hline

DenseNet121~\cite{huang2017densely} & 83.6$\pm$6.7 & 78.8$\pm$4.2    & 70.5$\pm$9.2        & 87.1$\pm$8.3        & 86.8$\pm$6.5      \\ \hline
DenseNet201~\cite{huang2017densely} & 85.6$\pm$4.9 & 80.5$\pm$2.7    & 72.4$\pm$7.1        & 88.6$\pm$2.1        & 89.9$\pm$3.2      \\ \hline
VGG11~\cite{simonyan2014very}       & \textbf{94.9$\pm$1.4} & \textbf{86.6$\pm$2.2}    & \textbf{74.7$\pm$4.7}        & \textbf{98.6$\pm$0.4}        & \textbf{98.1$\pm$0.4}      \\ \hline
VGG16~\cite{simonyan2014very}       & 93.6$\pm$1.3 & 85.9$\pm$1.6    & 73.8$\pm$4.4        & 97.0$\pm$0.5        & 98.01$\pm$0.6      \\ \hline
\end{tabular}
\end{table}

\subsection{Ablation Study}
The result with or without additional points from the correct reconstructions used for training are presented in Table~\ref{table:with_without}. The network used in this experiment are the one with the best performance in the cross validation experiment, \ie VGG11. Although with additional points for training, VGG11 has inferior performance regarding the classification of the POIs and their match points (86.6\% vs. 91.2\%), its performance on the other points in the correct reconstruction is consistent with the match points of POIs (98.2\% vs. 98.6\%), suggesting the network has a decent generalization capability. Such a stable and consistent performance demonstrates that the proposed framework can be ultilized in practice to provide guidance towards better neuron reconstruction.

\begin{table}[th]
\caption{The results w/o additional points from the correct reconstructions used for the training of VGG11. In each cell, the first number represents the average measurement, and the second number indicates the standard deviation. 'Specificity2' represents the result for the other points in the correct reconstructions which are not match points of POIs.
\label{table:with_without}}
\centering
\setlength{\tabcolsep}{3pt}
\begin{tabular}{|c|c|c|c|c|c|}
\hline
     & AUC (\%)  & Accuracy (\%) & Sensitivity (\%) & Specificity (\%) & Specificity2 (\%)\\ \hline
With       & 94.9$\pm$1.4 & 86.6$\pm$2.2    & 74.7$\pm$4.7        & 98.6$\pm$0.4          &98.2$\pm$0.2 \\ \hline
Without    & 96.0$\pm$1.3 & 91.2$\pm$1.7    & 90.1$\pm$4.5        & 92.2$\pm$1.5          &82.5$\pm$1.9  \\ \hline
\end{tabular}
\end{table}

\section{Conclusion}
In this study, we proposed a fully automatic framework for the quality control of neuron reconstruction. By formulating the problem into a binary classification task for each neuronal point and leveraging state-of-the-art deep learning technology, the proposed approach achieved a sensitive of 74.7\% for the test set, indicating its capability of detecting wrong tracing, and the high specificity of 98.6\% for the match points from the correct reconstructions suggests a low false alert rate. Furthermore, the network trained with a few additional points presented a consistent performance on all the other points in the correct reconstructions, which demonstrated the proposed approach can be used in practice to provide guidance for the annotators towards more accurate and efficient neuron reconstruction.

\clearpage
\bibliographystyle{splncs04}
\bibliography{main}
\end{document}